\pdfoutput=1
\documentclass{article}

\PassOptionsToPackage{table}{xcolor}
\PassOptionsToPackage{numbers,compress}{natbib}
\usepackage[preprint]{paper}

\usepackage[utf8]{inputenc}
\usepackage[T1]{fontenc}
\usepackage{url}

\usepackage{caption}
\usepackage{algorithm}
\usepackage{algorithmic}
\usepackage{algpseudocode}

\usepackage{microtype}
\usepackage{graphicx}
\usepackage{subcaption}
\usepackage{booktabs}

\usepackage{hyperref}

\usepackage{amsmath}
\usepackage{amssymb}
\usepackage{amsfonts}
\usepackage{mathtools}
\usepackage{amsthm}
\usepackage{nicefrac}

\usepackage[capitalize,noabbrev]{cleveref}

\usepackage{lipsum}
\usepackage{duckuments}
\usepackage{tcolorbox}
\tcbuselibrary{breakable}
\usepackage{wrapfig}
\usepackage{enumitem}
\usepackage{colortbl}
\usepackage{bm}
\usepackage{float}
\usepackage{multirow}
\usepackage{blindtext}
\usepackage{makecell}
\usepackage{adjustbox}
\usepackage{soul}
\usepackage{xspace}

\newcommand{\ie}{\textit{i}.\textit{e}., }
\newcommand{\eg}{\textit{e}.\textit{g}., }
\definecolor{mygreen}{rgb}{0, 0.6823, 0.7215}
\definecolor{GoogleRed}{RGB}{234, 67, 53}
\definecolor{GoogleBlue}{RGB}{66, 133, 244}
\definecolor{GoogleGreen}{RGB}{52, 168, 83}
\definecolor{JSViolet}{RGB}{71,15,244}
\definecolor{JSRed}{RGB}{205,44,78}
\definecolor{RowHighlight}{gray}{0.9}
\definecolor{cornellred}{rgb}{0.7, 0.11, 0.11}
\definecolor{diamondPurple}{RGB}{103, 58, 183}
\definecolor{myblue}{RGB}{47, 114, 173}
\definecolor{PINK}{RGB}{252, 81, 133}
\hypersetup{
  urlcolor   = PINK,
  linkcolor  = myblue,
  citecolor  = myblue,
  colorlinks = true,
}
\usepackage{pifont}
\newcommand{\cmark}{\textcolor{JSViolet}{\ding{51}}}
\newcommand{\xmark}{\textcolor{JSRed}{\ding{55}}}
\usepackage[normalem]{ulem}
\useunder{\uline}{\ul}{}

\DeclareUnicodeCharacter{221E}{\ensuremath{\infty}}

\crefformat{table}{Table~#2#1#3}
\crefformat{figure}{Figure~#2#1#3}
\crefformat{section}{Section~#2#1#3}
\crefformat{appendix}{Appendix~#2#1#3}
\crefformat{equation}{Eq.~#2(#1)#3}

\theoremstyle{plain}

\theoremstyle{definition}

\theoremstyle{remark}

\usepackage[textsize=tiny]{todonotes}

\newcommand{\mname}{IdleSpec\xspace}
\newcommand{\lname}{\textit{IdleSpec} (Exploiting \textbf{Idle} Time via \textbf{Spec}ulative Planning\xspace)}
\newcommand{\methodname}{IdleSpec\xspace}

\title{IdleSpec: Exploiting Idle Time via Speculative Planning for LLM Agents}

\newcommand\blfootnote[1]{%
  \begingroup
  \renewcommand\thefootnote{}\footnote{#1}%
  \addtocounter{footnote}{-1}%
  \endgroup
}

\author{%
  Daewon Choi$^{1}$ \quad Kyunghyun Park$^{1}$ \quad Woomin Song$^{1}$ \quad Saket Dingliwal$^{3\dagger}$ \\
  \bfseries Sai Muralidhar Jayanthi$^{2}$ \quad Jinwoo Shin$^{1}$ \quad Aram Galstyan$^{2}$ \\[4pt]
  \mdseries $^{1}$KAIST \qquad $^{2}$Amazon AGI \qquad $^{3}$Together AI
}

\begin{document}

\maketitle
\blfootnote{\ignorespaces$^{\dagger}$Work done at Amazon.}
\begin{abstract}
Large language model (LLM)-based agents solve complex tasks by leveraging multi-step reasoning with iterative tool calls and environment interactions, which incur idle time while waiting for observations. Despite the prevalence of idle time in most agentic scenarios, existing works treat it as an unavoidable overhead 
or propose restricted solutions that overlook varying computational budgets across different tool calls and future
observation uncertainty, thereby leading to suboptimal utilization of idle time. 
In this paper, we introduce \mname, a scalable and generic inference approach that leverages idle-time computation
 to improve agent performance while minimizing latency overhead. Specifically, \mname iteratively generates 
 plan candidates during idle periods and, once observations become available, aggregates them to guide the next reasoning step.
  For effective plan generation under observation uncertainty, \mname samples between complementary drafting strategies 
  (\ie progressive and recovery) from a learned distribution that is updated via posterior feedback. 
  Our experiments demonstrate that \mname significantly improves agent performance in various agentic scenarios 
  by effectively utilizing idle time. In particular, on the GAIA and FRAMES, \mname achieves 55.6\% average accuracy with Gemini-2.5-Flash, 
  surpassing the vanilla baseline without idle-time usage by 5.1\%. Furthermore, for MLE-Bench, which involves substantial delay from code executions, \mname achieves performance gains of up to 9.1\% on the Any Medal rate, highlighting its generalizability to long-horizon tasks.
\end{abstract}
\section{Introduction}
\label{sec:intro}
\begin{figure*}[!t]
\centering\small
\centering
\includegraphics[width=\linewidth,height=0.45\linewidth]{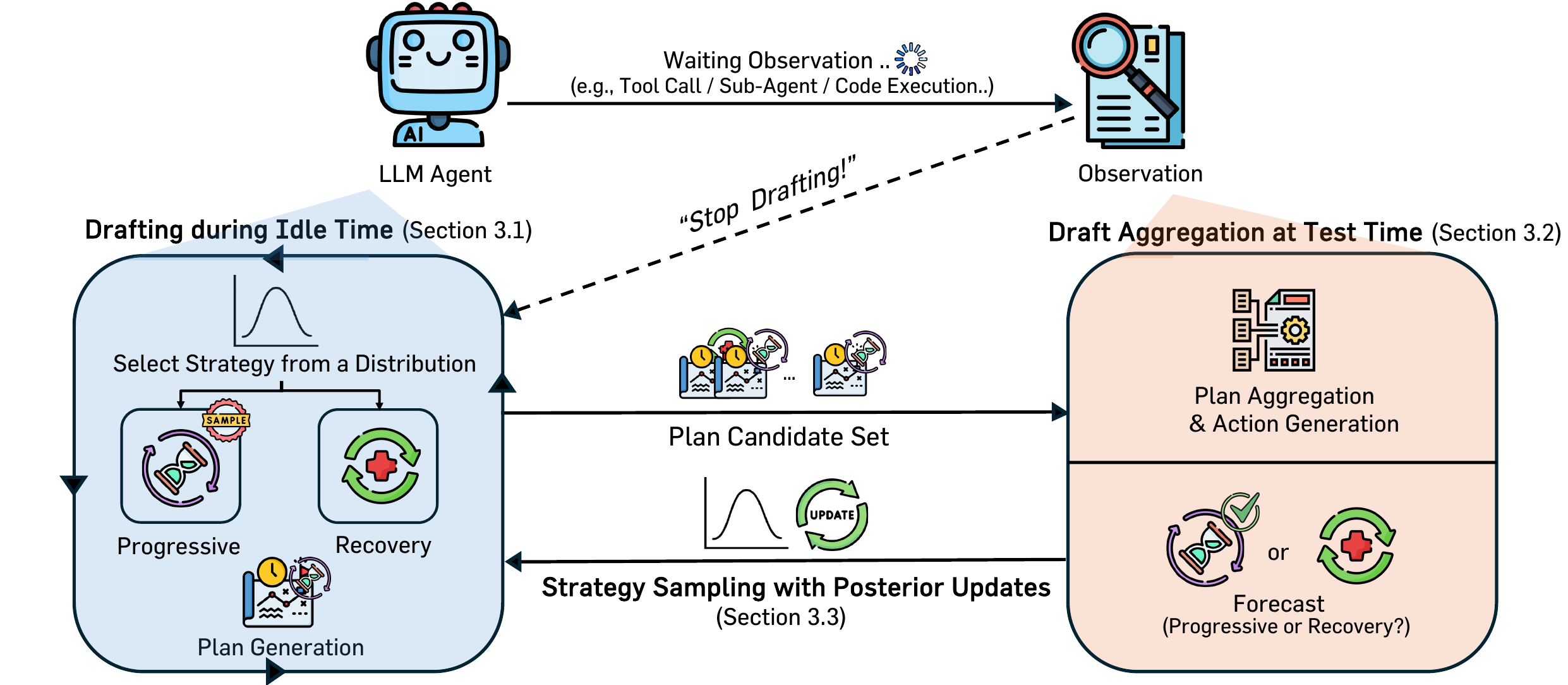}
\vspace{-2mm}
\caption{
\textbf{Overview of \mname.}
\textbf{(a) Idle-Time Drafting:} during tool execution, the agent iteratively drafts plan candidates by sampling between \textit{Progressive} and \textit{Recovery} strategies, and terminates drafting once the observation arrives.
\textbf{(b) Draft Aggregation:} the agent aggregates the candidates with the observation into a refined action and forecasts whether the trajectory is on track or requires recovery.
\textbf{(c) Posterior Update:} the forecast signal updates the strategy distribution, biasing future drafts toward strategies effective in the current context.
}
\label{fig:concept}
\vspace{-1em}
\end{figure*}

Large language model (LLM)-based agents \cite{yao2022react, sarukkai2025react2} have shown remarkable progress across diverse domains, including general problem solving \cite{mialon2023gaia, phan2025humanity}, web navigation \cite{zhou2023webarena, deng2023mind2web}, code generation \cite{jimenez2023swe, chan2024mle-bench}, and scientific research \cite{lu2024ai}, by actively leveraging external tools such as web search, code interpreters \cite{zhang2024codeagent}, and even other agents \cite{zhang2025agentastool, jiang2025agentadaptation}. Such agents commonly follow a multi-step reasoning loop \cite{yao2022react} in which the agent iteratively reasons, invokes tools, and incorporates the resulting observations, incurring frequent waiting time at each step, referred to as \textit{idle time}. Despite its prevalence in agent execution, the potential of idle time for scaling performance remains largely underexplored, as existing works mainly target efficiency (\eg asynchronously invoking independent tool calls \cite{gim2024async1}). \citet{lin2025sleeptime} first treat idle periods as an opportunity for additional computation that produces auxiliary context, but their design targets LLM–user interactions and remains suboptimal for agentic settings: it ignores variation in tool-call durations, utilizing only 13.7\% of the total idle time on GAIA (see Figure~\ref{fig:gaia-scatter}), and relies on the assumption of predictable future queries, which rarely holds when tool observations are inherently uncertain, often degrading performance (see \textit{Sleep-time Compute} in Table~\ref{tab:main1}).

This raises a key question:
\textit{How can we develop a robust, generalizable solution for exploiting idle time arising from agentic interactions to improve agent performance on complex tasks?
} This naturally leads us to analyze idle time across diverse agentic scenarios and explore effective strategies for leveraging it. Our analysis yields the following key observations: 

\begin{itemize}
\item \textit{Reasoning budgets for idle time are substantial but highly variable across tool calls.} 
We observe that the available reasoning budget during idle time is sufficiently large across diverse benchmarks---including tool-augmented reasoning (GAIA), multi-hop search (FRAMES), and execution-heavy environments (MLE-Bench)---providing additional computation to improve performance. However, the amount of available budget varies significantly across individual tool calls.

\item \textit{Planning is more effective than other agent strategies during idle time.} 
Among representative agent strategies, \ie summarization, reflection, and planning, we find that planning yields the most consistent performance improvements when generated during idle time.
\end{itemize}

Motivated by these findings, we propose \lname, a scalable inference framework that exploits idle time via speculative planning; see the overview in Figure~\ref{fig:concept}. Specifically, we adopt a simple yet effective two-phase strategy: (a) during idle time, the agent iteratively generates candidate plans; (b) once observations arrive, these candidates are aggregated and conditioned to guide subsequent reasoning. For higher utilization of idle time, the agent iteratively drafts candidates until an observation becomes available and terminates immediately upon observation arrival, thereby enabling higher utilization of idle time.

The key challenge of planning during idle time lies in handling observation uncertainty.
Since observations are not yet available during idle periods, generated plans may become invalid or suboptimal once the observation becomes available.
To mitigate this uncertainty, we design a drafting strategy distribution that dynamically samples between two complementary strategies: \textit{Progressive}, which assumes favorable observations and emphasizes exploitation, and \textit{Recovery}, which explores alternative solution paths assuming potential failure from observation. 
By generating both forward-progressing and recovery-oriented drafts through sampling, the agent improves coverage over plausible future observations.
Furthermore, this strategy distribution is updated via posterior feedback, whereby the agent evaluates its current progress and forecasts the most suitable drafting strategy for subsequent idle periods. This adaptive update enables the drafting behavior to be adjusted during inference, resulting in robust performance across diverse execution contexts.

To validate the effectiveness and generality of \mname, we conduct extensive experiments across three diverse agentic benchmarks: tool-augmented reasoning, multi-hop search, and long-horizon interactive tasks.
We first evaluate \mname on the GAIA benchmark \cite{mialon2023gaia}, which requires diverse capabilities including web search, file inspection, and code execution, and on FRAMES~\cite{krishna2024frames}, which requires multi-hop search via repeated agent-as-tool invocations.
We show that \mname consistently improves performance across various LLM backbones, \eg surpassing the vanilla baseline by 4.6\% and 6.8\% in average accuracy on Gemma4-E4B and Qwen3.5-4B, respectively. The performance gains are also more consistent than those of prior idle-time approaches \cite{lin2025sleeptime}, which exhibit lower utilization of total idle time and, in some cases, even degrade baseline performance. 
Next, we verify the generality of \mname on MLE-Bench \cite{chan2024mle-bench}, a benchmark that involves substantial idle time due to code execution for machine learning engineering scripts. In this setting, \mname significantly outperforms all baselines, highlighting its applicability to complex, long-horizon interactive tasks.

\textbf{Contributions.}
Our contributions are as follows:
\begin{itemize}
    \item We conduct a systematic analysis of idle time in LLM-based agents across diverse agentic scenarios, showing that idle periods are substantial but vary significantly across tool calls. We further establish that planning yields more effective use of idle-time computation than other agentic strategies such as summarization and reflection.
    
    \item We propose \mname, a scalable inference-time framework that exploits idle time in agentic interactions via speculative planning. 
    \mname iteratively and adaptively generates candidate plans by sampling from a strategy distribution, and aggregates them once the observations for the next step become available. 
    
    \item We validate \mname across three diverse agentic benchmarks, \ie GAIA, FRAMES, and MLE-Bench, spanning tool-augmented reasoning, multi-hop search, and long-horizon interactive tasks, and demonstrate consistent performance gains over existing baselines. 
\end{itemize}

\section{Related Works}
\label{sec:related-works}
\textbf{LLM Agents.}
Large language models (LLMs) have demonstrated strong reasoning capabilities through multi-step reasoning paradigms such as Chain-of-Thought (CoT)~\cite{wei2022chain}, enabling them to solve increasingly complex tasks.
Building on this, existing works~\cite{yao2022react, sarukkai2025react2, rawat2025pre, zhang2024codeagent} have considered LLMs as sequential decision-making policies, referred to as agents, that interact with external environments.
These approaches formalize agent execution as an iterative reason–act–observe loop, where the agent reasons about the next action, invokes external tools, and conditions on the resulting observations to guide subsequent steps.
Several studies~\cite{schick2023toolformer, qu2025toollearning, zhang2024codeagent} have further shown that LLMs can be augmented to invoke external tools such as web search, code execution, and databases, thereby extending their capabilities beyond training knowledge. More recently, the definition of tools has been generalized to include other agents, enabling task decomposition~\cite{fourney2024magentic, zhang2025agentorchestra}, and intermediate result verification~\cite{lifshitz2025multi}.
For example, recent works~\cite{zhang2025agentastool, jiang2025agentadaptation} treat agents themselves as callable tools, allowing LLM agents to coordinate with specialized sub-agents on a per-task basis.
Despite these advances in tool usage within agentic workflows, most existing approaches treat the time spent waiting for observations as an unavoidable delay.
In this work, we view this waiting period as an opportunity for additional computation to improve performance and propose a novel inference-time method called \mname.

\textbf{Idle Time in LLM Agents.}
As LLM agents interact with tools and external environments, execution time is often dominated not by LLM inference itself, but by tool execution and environment response delays.
To leverage this delay, most existing works~\cite{gim2024async1, li2025continuum, biswas2026sutradhara} focus on improving system efficiency by overlapping independent computations or through system-level execution management.
For instance, \citet{gim2024async1} and \citet{biswas2026sutradhara} explore asynchronous and non-blocking function-calling mechanisms that allow LLM inference to proceed without waiting for tool execution to complete, while \citet{li2025continuum} observes that frequent tool invocations incur significant overhead due to cache eviction and rescheduling between turns, and proposes cache-aware scheduling.
A parallel line of work targets latency reduction through speculation: \citet{ye2026speculative} and \citet{nichols2025optimizing} introduce predict-and-verify paradigms in which a fast speculator executes likely-correct actions or tool calls that a slower target model verifies, while \citet{hua2025interactive} applies the same two-model speculation pattern to multi-step planning.
These methods can be viewed as single-mode speculation: they assume the current trajectory will succeed and commit to a single speculative continuation along that path.
In contrast, \mname{} models the inherent uncertainty of future observations through complementary progressive and recovery drafts, and exploits idle time to improve task performance rather than to amortize already-required computation.
Recent work on \textit{Sleep-Time Compute}~\cite{lin2025sleeptime} partially addresses this performance gap by treating idle periods as opportunities for pre-computation that generates auxiliary context. However, it does not explicitly account for the highly variable idle-time budgets in agent executions and relies on the restrictive assumption that future query patterns are predictable, which rarely holds in agentic scenarios where tool-call observations are inherently uncertain.
In this paper, we propose a robust solution that effectively exploits idle time to improve agent performance and can be seamlessly applied across diverse, realistic agentic scenarios.

\begin{figure}[t]
\centering\small
\begin{subfigure}[t]{0.32\linewidth}
  \centering
  \includegraphics[width=\linewidth]{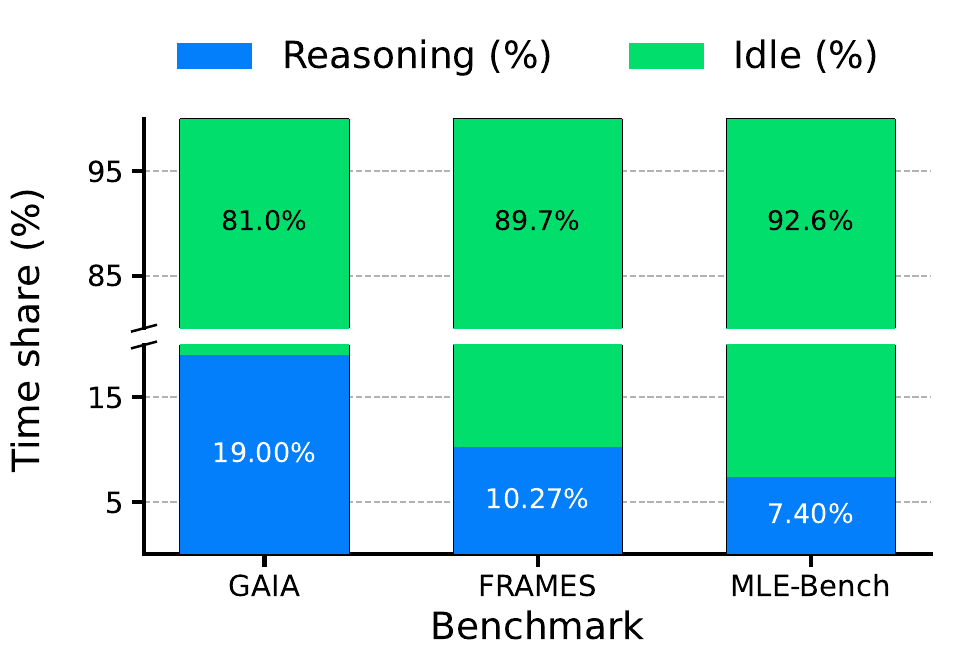}
  \caption{Idle vs.\ Reasoning Share}
  \label{fig:observation-source}
\end{subfigure}\hfill
\begin{subfigure}[t]{0.32\linewidth}
  \centering
  \includegraphics[width=\linewidth]{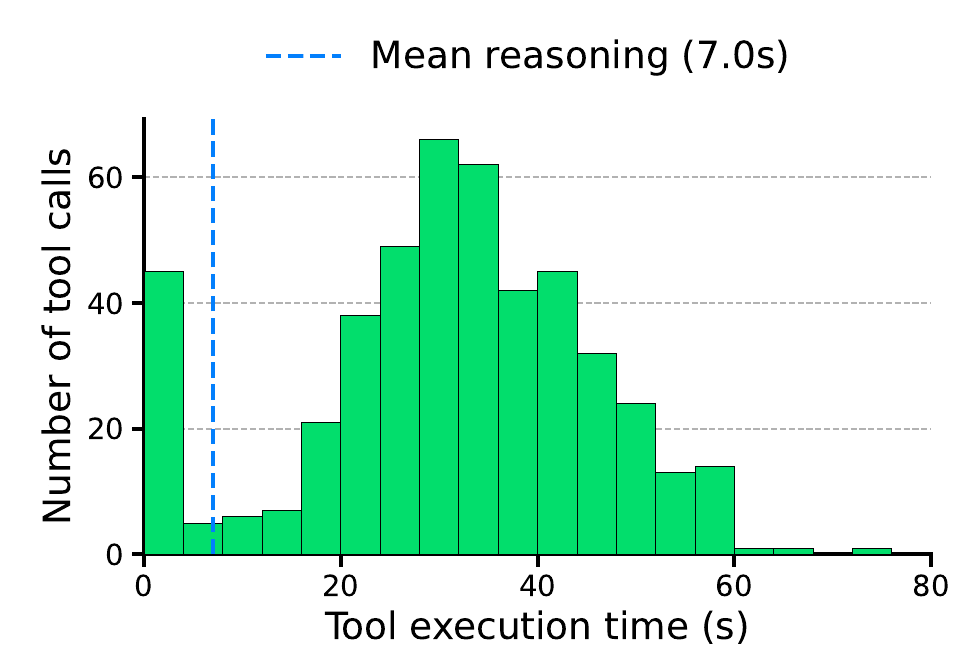}
  \caption{Tool Execution Time}
  \label{fig:observation-tool-execution-time}
\end{subfigure}\hfill
\begin{subfigure}[t]{0.32\linewidth}
  \centering
  \includegraphics[width=\linewidth]{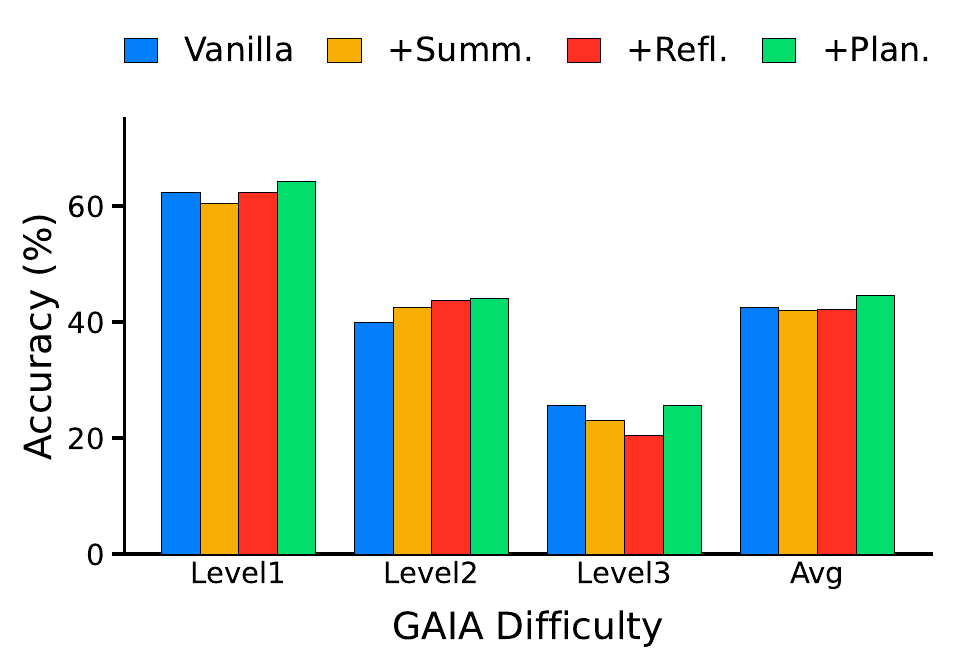}
  \caption{Idle-Time Strategies}
  \label{fig:observation-methods}
\end{subfigure}
\caption{\textbf{How Can We Leverage Idle Time in LLM Agents?}
(a) Reasoning time vs.\ tool execution time (\ie idle time) across benchmarks.
(b) Histogram of per-call tool execution times. 
(c) Accuracy of three idle-time strategies (Summarization, Reflection, Planning) vs.\ vanilla.}
\label{fig:observation-idle}
\end{figure}

\section{How Can We Leverage Idle Time in LLM Agents?}\label{sec:observation}
In this section, we investigate the following key questions: how much reasoning budget is available during idle time, and which strategies are effective for utilizing it. To this end, we conduct experiments on 
three benchmarks:
(i) GAIA, in which agents invoke diverse external tools such as web 
search, file readers, and multimodal parsers;
(ii) FRAMES, which contains multi-hop questions requiring long chain
of sequential search-based tool calls; and
(iii) MLE-Bench, which involves ML workloads (model training, 
evaluation) that produce substantial environment response delays.

\subsection{Source of Idle Time}
To quantify the reasoning budget available during idle time, we visualize
the ratio of tool execution to reasoning time. As shown in
Figure~\ref{fig:observation-source}, tool execution
dominates the total execution time (e.g., it is more than 12.5x larger
than reasoning time in MLE-Bench), leaving a substantial reasoning budget
unused.
Beyond this aggregate gap, the per-call statistics in Figure~\ref{fig:observation-tool-execution-time} reveal that idle durations are highly heterogeneous within a single trajectory, ranging from shorter than a single reasoning step (the dashed line) to more than 10x longer. The long tail of this distribution accounts for the majority of total idle time and is large enough to support \emph{multiple} rounds of speculative computation, making it the opportunity for idle-time exploitation.

\subsection{Strategies for Idle Time}\label{subsec:obs2}
To identify effective idle-time strategies, we compare three representatives against a vanilla baseline without idle-time computation:
(i) \textit{summarization}, which compresses the interaction history;
(ii) \textit{reflection}, which evaluates the current action based on past trajectories; and
(iii) \textit{planning}, which generates candidate plans conditioned on plausible future observations.
As shown in Figure~\ref{fig:observation-methods}, only planning consistently matches or exceeds the baseline across all difficulty levels, achieving the highest average accuracy, while summarization and reflection fail to deliver consistent gains and even degrade performance (e.g., a notable accuracy drop for reflection on GAIA Level 3).
We attribute this to differences in how each method handles observation uncertainty during idle time. Summarization and reflection must commit to an incomplete interpretation of the trajectory before future observations arrive (e.g., reflection misjudging a valid action as erroneous), which then propagates into subsequent decisions. Planning avoids this by admitting a conditional formulation over possible future observations (e.g., if A holds, execute X; otherwise Y), making idle-time compute resilient to this uncertainty.

\section{IdleSpec: Exploiting Idle Time via Speculative Planning for LLM Agents}
\label{sec:method}
In this section, we propose IdleSpec, a novel inference framework that exploits idle time in LLM agents via speculative planning. We design a two-phase strategy: it drafts two complementary types (\ie \textit{progressive} and \textit{recovery}) of plan candidates during idle periods (Section~\ref{subsec:drafting}), and aggregates them once observations arrive and uses them for subsequent reasoning (Section~\ref{subsec:verification}). To ensure effective drafting under observation uncertainty, we propose adaptive drafting strategy sampling with posterior updates (Section~\ref{subsec:adaptive}). The overall framework is illustrated in Figure~\ref{fig:concept}.
\subsection{Drafting during Idle Time}
\label{subsec:drafting}

\paragraph{Progressive Drafting.}
The agent assumes that the forthcoming observation will be successfully obtained and will contribute to task success. Under this assumption, the agent proactively drafts the next plan to be executed once the observation becomes available, 
focusing on forward progress.

\begin{tcolorbox}[title=Progressive Drafting Prompt, fontupper=\small, width=\linewidth]
You are an expert AI assistant tasked with generating the most effective and efficient NEXT ACTION STEP to take after the current action step completes and its observation becomes available.

The current action step is running in parallel.
You will not see the observation yet; assume it will arrive and will be used to decide the next step.
\end{tcolorbox}

\paragraph{Recovery Drafting.}
The agent prepares for the case in which the forthcoming observation shows no progress (e.g., the tool returns an empty or off-target result). Under this assumption, the agent drafts an alternative plan that pursues the same sub-goal from a different angle.

\begin{tcolorbox}[title=Recovery Drafting Prompt, fontupper=\small, width=\linewidth]
You are an expert AI assistant tasked with generating a RECOVERY strategy while the current action step is executing.

Assume that the current action step fails to make progress or gets stuck.
Propose EXACTLY one distinct recovery plan that takes a different approach from what has already been tried.
\end{tcolorbox}

\paragraph{Idle-aware Iterative Drafting.}
As shown in Figure~\ref{fig:observation-source}, idle time varies across tools and environments,
making the appropriate number of plan candidates highly variable.
To address this, we design drafting as an iterative procedure that runs concurrently with tool execution and terminates immediately once the tool response becomes available.
While waiting for the observation, the agent repeatedly generates plan candidates at each iteration.
These drafts are accumulated over time and fed back into the prompt at each iteration, allowing the model to condition on previously generated plans.
This design enables the agent to generate diverse plan candidates, while reducing unnecessary drafting cost. Once the tool execution completes, drafting halts and the accumulated candidates are finalized.
As a result, the final candidate set $\mathcal{D}_\text{final}$ is given by:
\begin{equation}
\mathcal{D}_\text{final} = \mathcal{D}_\text{prog} \cup \mathcal{D}_\text{rec}, 
\end{equation}
where $\mathcal{D}_\text{prog}$ is the candidate set from progressive drafting and $\mathcal{D}_\text{rec}$ is from recovery drafting.

\subsection{Draft Aggregation at Test Time}
\label{subsec:verification}

Once the observation becomes available, the agent proceeds to the next reasoning step by aggregating the candidate set $\mathcal{D}_\text{final}$ for guidance.

\textbf{Aggregation.}
Given the plan candidate set $\mathcal{D}_\text{final}$, the agent aggregates the generated drafts by conditioning its next reasoning step on the entire set. 
Importantly, the agent is not required to strictly follow any individual candidate; the plans are treated as \textit{reference points} that may guide the next reasoning step. 
This careful design is crucial because drafts generated during idle time are produced under observation uncertainty; we observe that directly forcing such raw plans can lead to suboptimal performance (see Table~\ref{tab:aggregation} in Section~\ref{sec:ablation}).

\begin{tcolorbox}[title=Aggregation Prompt, fontupper=\small, width=\linewidth]
The following are candidate plans generated for the next step:

\textbf{Plan 1:} \textbf{\{candidate plan 1\}} \\[2pt]
\textbf{Plan 2:} \textbf{\{candidate plan 2\}} \\[2pt]
$\vdots$ \\[2pt]
\textbf{Plan N:} \textbf{\{candidate plan N\}}

Based on the observation, you may use these plans as starting points, but you are free to synthesize a different or improved next step if none of them fit well.
\end{tcolorbox}

\subsection{Strategy Sampling with Posterior Updates}
\label{subsec:adaptive}

We now describe how the agent decides which drafting strategy (i.e., progressive and recovery) to use at each idle-time iteration. 
Effective drafting under observation uncertainty requires covering a diverse set of plausible future situations, while still concentrating computation on the strategy most likely to be useful for the current trajectory. 
To balance these two objectives, we propose a simple yet effective strategy-selection mechanism based on Thompson sampling~\cite{shipra2012thompson}: we derive a forecast signal from each observation and use posterior sampling to adaptively select the strategies.

\paragraph{Forecast.}
At the end of each step, once the observation is available, the agent produces a forecast for the subsequent idle period. The forecast outputs a binary signal $\ell \in \{\textsc{Prog}, \textsc{Rec}\}$, 
indicating which drafting strategy is more promising: progressive (continuing along the current trajectory) or recovery (mitigating potential future failures). 
The forecast prompt is provided in Appendix~\ref{sec:appendix_prompt_templates} (Figure~\ref{fig:prompt_forecast}).

\paragraph{Probabilistic Model.}
Let $p \in [0,1]$ denote the probability that the forecast favors the progressive strategy,
\begin{equation}
p = \Pr(\ell = \textsc{Prog}),
\end{equation}
and place a Beta prior over $p$,
\begin{equation}
p \sim \mathrm{Beta}(\alpha, \beta),
\end{equation}
where $\alpha$ and $\beta$ count past forecast signals corresponding to \textsc{Prog} and \textsc{Rec}, respectively. 
We initialize $\alpha = \beta = 1$ to reflect a uniform prior in the absence of any forecast feedback. After each forecast signal $\ell$, we update the posterior by incrementing the count corresponding to $\ell$:
\begin{equation}
(\alpha, \beta) \leftarrow
\begin{cases}
(\alpha + 1, \beta), & \text{if } \ell = \textsc{Prog}, \\
(\alpha, \beta + 1), & \text{if } \ell = \textsc{Rec}.
\end{cases}
\label{eq:posterior-update}
\end{equation}

\paragraph{Adaptive Strategy Sampling.}
At each drafting iteration, we draw a preference value $\hat{p} \sim \mathrm{Beta}(\alpha, \beta)$ and select the drafting strategy as $\mathbf{s} = \textsc{Prog}$ if $\hat{p} > 0.5$ and $\mathbf{s} = \textsc{Rec}$ otherwise. The posterior is held fixed throughout the upcoming idle period and updated again once the next observation arrives.
This procedure naturally allocates more idle-time computation to the strategy that the forecast favors, while retaining stochastic exploration to handle uncertainty and preference shifts during execution.

\section{Experiments}
\label{sec:experiments}
{\renewcommand{\arraystretch}{1.05}
\begin{table}[t]
\centering
\small
\caption{\textbf{Results on General Agent Benchmarks.} We report accuracy (\%) on GAIA and FRAMES. \textbf{Bold} and \underline{underline} indicate best and runner-up results, respectively.}\label{tab:main1}
\vspace{1em}
\resizebox{0.96\linewidth}{!}{
\begin{tabular}{lcccccc}
    \toprule
    Method & Use idle? & GAIA Level1 & GAIA Level2 & GAIA Level3 & FRAMES & Average\\
    \midrule
    \multicolumn{7}{c}{\cellcolor[HTML]{E0E0E0}\textit{Gemma4-E4B}}\\
    \midrule
    Vanilla              & \xmark & 36.5$_{\pm 3.6}$ & 25.6$_{\pm 0.9}$ & \underline{11.5}$_{\pm 0.0}$ & 50.0$_{\pm 5.3}$ & 30.9\\
    +Sequential Revision & \xmark & 39.6$_{\pm 1.5}$ & \underline{26.4}$_{\pm 1.5}$ & 9.0$_{\pm 4.8}$ & \underline{53.3}$_{\pm 7.0}$ & \underline{32.1}\\
    +Sleep-time Compute  & \cmark & \underline{41.5}$_{\pm 4.1}$ & 23.6$_{\pm 3.8}$ & 9.0$_{\pm 4.8}$ & 50.0$_{\pm 5.3}$ & 31.0\\
    \midrule
    \rowcolor{green!10}
    \textbf{+ \methodname (Ours)} & \cmark & \textbf{43.4}$_{\pm 2.7}$ & \textbf{28.3}$_{\pm 2.1}$ & \textbf{14.1}$_{\pm 3.5}$ & \textbf{56.0}$_{\pm 3.5}$ & \textbf{35.5}\\
    \midrule
    \multicolumn{7}{c}{\cellcolor[HTML]{E0E0E0}\textit{Qwen3.5-4B}}\\
    \midrule
    Vanilla              & \xmark & 37.7$_{\pm 7.6}$ & 26.4$_{\pm 2.7}$ & \underline{11.5}$_{\pm 7.7}$ & \underline{57.3}$_{\pm 6.4}$ & 33.2\\
    \midrule
    +Sequential Revision & \xmark & \textbf{41.5}$_{\pm 5.0}$ & \underline{27.9}$_{\pm 3.1}$ & \underline{11.5}$_{\pm 7.7}$ & 56.0$_{\pm 4.0}$ & \underline{34.2}\\
    +Sleep-time Compute  & \cmark & 36.5$_{\pm 2.9}$ & 21.3$_{\pm 4.7}$ & 9.0$_{\pm 4.4}$ & 50.7$_{\pm 1.2}$ & 29.4\\
    \rowcolor{green!10}
    \textbf{+ \methodname (Ours)} & \cmark & \underline{40.9}$_{\pm 3.9}$ & \textbf{31.8}$_{\pm 3.6}$ & \textbf{21.8}$_{\pm 5.9}$ & \textbf{65.3}$_{\pm 4.2}$ & \textbf{40.0}\\
    \midrule
    \multicolumn{7}{c}{\cellcolor[HTML]{E0E0E0}\textit{Gemini-2.5-Flash}}\\
    \midrule
    Vanilla              & \xmark & 62.3$_{\pm 2.0}$ & 39.9$_{\pm 3.6}$ & \underline{25.6}$_{\pm 9.7}$ & 74.0$_{\pm 2.0}$ & 50.5\\
    \midrule
    +Sequential Revision & \xmark & 61.6$_{\pm 3.9}$ & 42.3$_{\pm 1.8}$ & 21.8$_{\pm 4.4}$ & \underline{76.7}$_{\pm 1.2}$ & \underline{50.6}\\
    +Sleep-time Compute  & \cmark & \underline{62.9}$_{\pm 2.9}$ & \underline{44.2}$_{\pm 2.0}$ & 20.5$_{\pm 5.9}$ & 72.7$_{\pm 1.2}$ & 50.1\\
    \rowcolor{green!10}
    \textbf{+ \methodname (Ours)} & \cmark & \textbf{66.0}$_{\pm 3.8}$ & \textbf{46.5}$_{\pm 4.2}$ & \textbf{32.1}$_{\pm 4.4}$ & \textbf{78.0}$_{\pm 4.0}$ & \textbf{55.6}\\
    \bottomrule
\end{tabular}}
\end{table}}
{\renewcommand{\arraystretch}{1.25}
\begin{table*}[t]
\centering
\small
\caption{\textbf{Results on MLE-Bench Lite.} 
We report the percentage of tasks that yield a submission, a valid submission, 
an above-median score, and each Kaggle medal tier (Bronze/Silver/Gold).
All methods are evaluated under the standard 24-hour wall-clock budget per task, using Gemini-2.5-Flash as the backbone. 
\textbf{Bold} and \underline{underline} indicate best and runner-up results, respectively.}
\label{tab:main2}
\resizebox{1.0\textwidth}{!}{
\begin{tabular}{lcccc cccc}
    \toprule
    \multirow{2}{*}{Method} & \multirow{2}{*}{Use idle?} &
    \multirow{2}{*}{\makecell{Made\\Submission\\(\%)}} &
    \multirow{2}{*}{\makecell{Valid\\Submission\\(\%)}} &
    \multirow{2}{*}{\makecell{Above\\Median\\(\%)}} &
    \multicolumn{4}{c}{Any Medal (\%)} \\
    \cmidrule(lr){6-9}
    & & & & & Bronze & Silver & Gold & Total \\
    \midrule
    Vanilla & \xmark &
    86.4 & 77.3 & 40.1 & 9.1 & 9.1 & 18.2 & 36.4 \\
    \midrule
    +Sequential Revision & \xmark &
    \underline{90.1} & \underline{81.8} & \underline{50.0} & 13.6 & 9.1 & 18.2 & \underline{40.9} \\

    +Sleep-time Compute & \cmark &
    \underline{90.1} & \underline{81.8} & \textbf{59.1} & 13.6 & 9.1 & 13.6 & 36.4 \\
    \rowcolor{green!10}
    \textbf{+ \methodname (Ours)} & \cmark &
    \textbf{95.5} & \textbf{86.4} & \textbf{59.1} & 13.6 & 13.6 & 18.2 & \textbf{45.5} \\
    \bottomrule
\end{tabular}
}
\end{table*}
}

\begin{figure*}[t]
\centering
\begin{minipage}[c]{0.42\linewidth}
\centering
\includegraphics[width=\linewidth]{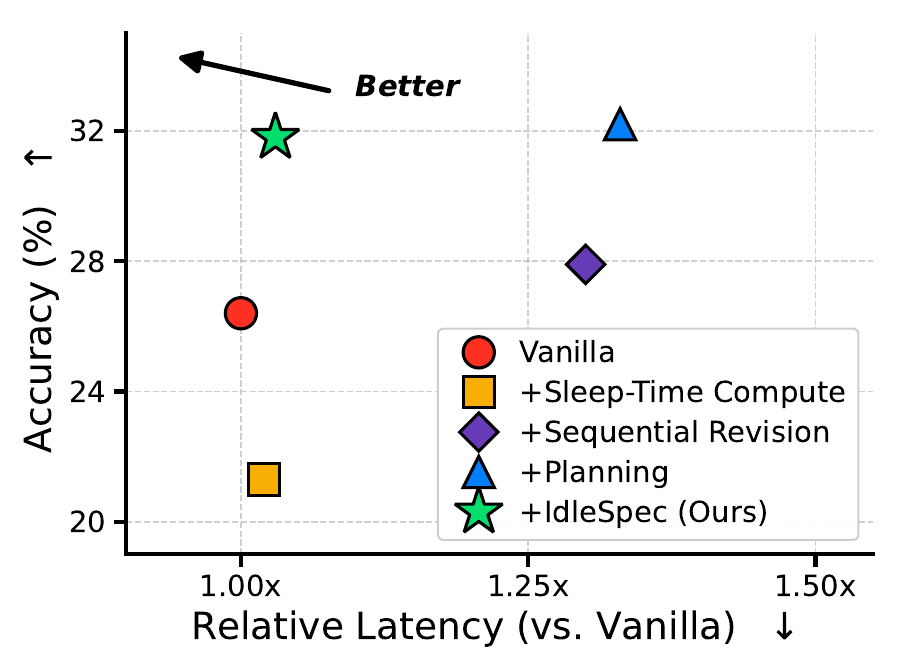}
\captionof{figure}{\textbf{Latency--Accuracy Trade-off.} All measurements were performed on vLLM using an NVIDIA A6000 GPU.}
\label{fig:latency-vs-acc}
\end{minipage}\hfill
\begin{minipage}[c]{0.55\linewidth}
\centering
\captionof{table}{\textbf{Idle-Time vs.\ Test-Time Token Usage.} Accuracy (\%) and output token usage during idle time and test time on GAIA Level 2 using Qwen3.5-4B.}
\label{tab:token-analysis}
\resizebox{\linewidth}{!}{
\begin{tabular}{lcc}
\toprule
Method & Accuracy & Tokens (Idle / Test)\\
\midrule
Vanilla              & 26.4 & (- / 7126)\\
\midrule
+Sequential Revision &  27.9 & (- / 11354)\\
+Planning        &  32.2 & (- / 12234)\\
+Sleep-time Compute & 21.3 & (3393 / 7048)\\
\rowcolor{green!10}
\textbf{+ \methodname (Ours)} & 31.8 & (5284 / 5966)\\
\bottomrule
\end{tabular}}
\end{minipage}
\end{figure*}

We design experiments to answer the following questions:
\begin{itemize}[leftmargin=*,itemsep=0mm]
\item \textbf{RQ1:} Does \mname improve task performance on general agentic benchmarks (GAIA, FRAMES) across diverse LLM backbones? (Table~\ref{tab:main1} in Section~\ref{sec:main-results})
\item \textbf{RQ2:} Does \mname generalize to execution-heavy environments with substantial tool-execution times (MLE-Bench)? (Table~\ref{tab:main2} in Section~\ref{sec:main-results})
\item \textbf{RQ3:} How efficiently does \mname utilize idle time, is it compatible with other test-time scaling methods, and which design choices matter? (Figure~\ref{fig:latency-vs-acc} and Tables~\ref{tab:token-analysis}, \ref{tab:planning_composition}, \ref{tab:ultra-short-impact}, \ref{tab:ablation} in Section~\ref{sec:ablation})
\end{itemize}

\textbf{Datasets.}
We evaluate \mname on three benchmarks: GAIA~\cite{mialon2023gaia} (general agent, full 165-task validation split), FRAMES~\cite{krishna2024frames} (multi-hop QA, 50 samples), 
and MLE-Bench Lite~\cite{chan2024mle-bench} (ML engineering, 24-hour budget per task).

\textbf{Models.}
Our experiments primarily use the proprietary model Gemini-2.5-Flash.
In addition, we consider Gemma4-E4B and Qwen3.5-4B to validate \mname with recent small open-source models.

\textbf{Baselines.}
To evaluate the effect of idle-time utilization, we include a vanilla
baseline that performs no additional computation during idle time. Additionally, we include Sequential Revision \cite{zhu2025scaling}, which performs revision step after each new observation.
To compare against approaches that utilize idle time, we consider Sleep-Time Compute \cite{lin2025sleeptime}, which generates auxiliary contexts during idle periods via naive prompting and reuses them at test time.\footnote{More details including implementation details, evaluation protocols are provided in Appendix~\ref{sec:appendix_details}.}

\subsection{Main Results}\label{sec:main-results}

\textbf{Results on General Agent Benchmarks.}
We evaluate \methodname{} on the GAIA and FRAMES benchmarks, which involve a diverse tool suite (e.g., text and audio processing, search agents) to solve complex multi-hop reasoning problems.
As shown in Table~\ref{tab:main1}, \methodname{} consistently improves task success rates over all baselines across both benchmarks and across all three GAIA difficulty levels. \methodname{} achieves the highest average accuracy on every backbone, improving over Vanilla by 4.6\% on Gemma4-E4B (30.9 $\rightarrow$ 35.5), 6.8\% on Qwen3.5-4B (33.2 
$\rightarrow$ 40.0), and 5.1\% on Gemini-2.5-Flash (50.5
$\rightarrow$ 55.6). The gains are most pronounced on the harder Levels~2 and~3, suggesting that idle-time speculation is most beneficial when the underlying reasoning task is difficult.
In contrast, Sleep-Time Compute exhibits limited or inconsistent improvements and, in some cases, even degrades performance. In particular, this degradation is especially pronounced with smaller open-source models such as Qwen3.5-4B: because Sleep-Time Compute freely pre-generates auxiliary context based on assumed future queries, hallucinations introduced at this stage mislead subsequent reasoning. \methodname{} avoids this failure mode through dual drafting that covers both progressive and recovery scenarios, guiding the agent toward correct trajectories. We provide qualitative examples of both behaviors in Appendix~\ref{sec:appendix_qualitative}.

\textbf{Results on MLE-Bench.} We further evaluate \methodname{} on MLE-Bench, which features long-horizon interactions and execution-heavy steps involving repeated code execution and model training. As shown in Table~\ref{tab:main2}, Sleep-Time Compute matches \methodname{} on the Above-Median metric but lowers the Gold-medal rate relative to Vanilla (13.6\% vs.\ 18.2\%) and yields no improvement in total medals --- consistent with our earlier observation that naively pre-generated context can mislead reasoning under observation uncertainty. In contrast, \methodname{} achieves the best results across nearly all metrics, raising the submission rate from 86.4\% to 95.5\%, the valid submission rate from 77.3\% to 86.4\%, and the overall medal rate from 36.4\% to 45.5\%, with consistent gains across Bronze, Silver, and Gold tiers.
Because MLE-Bench involves long-running tool calls such as model training, a non-trivial portion of the 24-hour wall-clock budget is otherwise spent waiting. \methodname{} converts this wasted idle time into useful LLM computation, generating diverse solution paths that translate into more valid submissions and more medal-tier solutions within the same time budget, highlighting that \methodname{} generalizes to execution-heavy agentic workloads.

\subsection{More Analysis}\label{sec:ablation}

\textbf{Efficiency Analysis.}
We analyze the efficiency of \methodname{} by comparing it against the original baselines (Sequential Revision and Sleep-Time Compute) and an additional test-time scaling baseline, \emph{Planning}, which generates a new plan after each tool-call observation before the next step.
As shown in Figure~\ref{fig:latency-vs-acc}, Sequential Revision and Planning improve accuracy by spending more tokens at test time, but this directly inflates end-to-end latency: Sequential Revision and Planning are about $1.30\times$ and $1.33\times$ slower than Vanilla, respectively.
\methodname{}, in contrast, achieves a Pareto improvement: it outperforms Sequential Revision and matches Planning's accuracy while keeping latency close to Vanilla, as most of its additional token computation is carried out during idle periods (Table~\ref{tab:token-analysis}).

\begin{table}[t]
\centering
\small
\setlength{\tabcolsep}{4pt}
\begin{minipage}[t]{0.45\linewidth}
\centering
\caption{\textbf{Compatibility with Test-Time Scaling.} Accuracy (\%) on GAIA Level~2 with Qwen3.5-4B. \methodname yields consistent gains when combined with existing test-time scaling methods. }
\label{tab:planning_composition}
\vspace{0.5em}
\begin{tabular}{lcc}
    \toprule
    Method & Use Idle? & Accuracy \\
    \midrule
    Sequential Revision  & \xmark & 27.9 \\
    \rowcolor{green!10}
    + \textbf{\methodname}  & \cmark & \textbf{32.2} \\
    \midrule
    Planning & \xmark & 32.2 \\
    \rowcolor{green!10}
    + \textbf{\methodname}  & \cmark  & \textbf{35.3} \\
    \bottomrule
\end{tabular}
\end{minipage}\hfill
\begin{minipage}[t]{0.45\linewidth}
\centering
\caption{\textbf{Effect of Idle-Time Length on Accuracy (\%).} GAIA samples are binned by their ultra-short ratio (the fraction of tool calls shorter than the reasoning step) with Gemini-2.5-Flash. Idle-time availability increases from High to Low.}
\label{tab:ultra-short-impact}
\vspace{1em}
\begin{tabular}{lccc}
    \toprule
    Method & High  & Medium  & Low \\
    \midrule
    Vanilla         & 50.0 & 39.0 & 39.0 \\
    \rowcolor{green!10}
    + \textbf{\mname} & 50.0 & \textbf{45.0} & \textbf{46.0} \\
    \bottomrule
\end{tabular}
\end{minipage}
\end{table}

\textbf{Compatibility with Test-Time Scaling.}
We investigate whether \methodname{} is complementary to existing test-time scaling methods, under the hypothesis that the two operate along orthogonal axes and can be seamlessly combined: \methodname{} scales computation along the \emph{idle-time} axis, while existing test-time scaling methods scale along the \emph{test-time} axis.
Since \methodname{} leaves the test-time procedure itself untouched, it can be layered on top of any such method without modification.
To verify this, we combine \methodname{} with the two test-time scaling baselines introduced above Sequential Revision and Planning,  while leaving each baseline's test-time procedure unchanged. As shown in Table~\ref{tab:planning_composition}, \methodname{} yields consistent gains on top of both baselines, improving Sequential Revision from 27.9\% to 32.2\% and Planning from 32.2\% to 35.3\%.
This gain likely arises because the progressive and recovery drafts generated during idle periods pre-expand the search space available at test time, which the test-time scaling method leverages to produce better-informed revisions and plan updates.

\textbf{Effect of Idle-Time Length.}
We analyze how the available idle time per step affects \methodname{}.
We define the \emph{ultra-short ratio} of a sample as the fraction of its tool calls whose execution time is shorter than a single LLM reasoning step. On GAIA, this ratio averages 25--27\% across the three difficulty levels (Appendix~\ref{sec:appendix_idle_length}, Table~\ref{tab:ultra-short-ratio}), indicating that most tool calls leave enough idle time for speculative drafting. Binning GAIA samples by ultra-short ratio into High (>0.75, least idle time), Medium (0.25--0.75), and Low (<0.25, most idle time) groups (Table~\ref{tab:ultra-short-impact}), we find that \methodname{} matches the vanilla baseline on the High group, where little usable idle time is available,  but delivers gains of 6.0 \% and 7.0 \% on the Medium and Low groups, respectively.

\begin{table}[t]
\centering
\small
\caption{\textbf{Ablation Studies.} Accuracy (\%) on GAIA and FRAMES when selectively ablating individual components of \methodname. (a) Drafting strategy on GAIA Level 1--3 using Gemini-2.5-Flash; (b, c) aggregation and selection strategies on FRAMES using Qwen3.5-4B.}
\label{tab:ablation}

\begin{subtable}[t]{0.40\linewidth}
\centering
\caption{Drafting Strategy}\label{tab:strategy}
\small
\setlength{\tabcolsep}{4pt}
\begin{tabular}{cccccc}
    \toprule
    Prog. & Rec. & L1 & L2 & L3 & Avg.\\
    \midrule
    \cmark & \xmark & 62.9 & 44.2 & 26.9 & 44.7\\
    \xmark & \cmark & 60.4 & 43.4 & 28.2 & 44.0\\
    \rowcolor{green!10}
    \cmark & \cmark & \textbf{66.0} & \textbf{46.5} & \textbf{32.1} & \textbf{48.2}\\
    \bottomrule
\end{tabular}
\end{subtable}\hspace{0.5em}
\begin{subtable}[t]{0.24\linewidth}
\centering
\caption{Aggregation}\label{tab:aggregation}
\small
\setlength{\tabcolsep}{4pt}
\begin{tabular}{lc}
    \toprule
    Method & Acc. \\
    \midrule
    Best-of-N & 59.3 \\
    Mandatory & 57.3 \\
    \rowcolor{green!10}
    \textbf{Ref. (Ours)} & \textbf{65.3} \\
    \bottomrule
\end{tabular}
\end{subtable}\hspace{0.5em}
\begin{subtable}[t]{0.24\linewidth}
\centering
\caption{Selection}\label{tab:selection}
\small
\setlength{\tabcolsep}{4pt}
\begin{tabular}{lc}
    \toprule
    Method & Acc. \\
    \midrule
    Random & 63.3 \\
    Forecast-Direct & 60.7 \\
    \rowcolor{green!10}
    \textbf{Adapt. (Ours)} & \textbf{65.3} \\
    \bottomrule
\end{tabular}
\end{subtable}
\end{table}

\textbf{Ablation Study.}
We ablate the three key design choices of \methodname{} --- drafting strategy, aggregation scheme, and selection rule.
First, Table~\ref{tab:strategy} shows that using only progressive or only recovery drafts underperforms, with recovery-only being especially harmful as the agent keeps switching plans without sufficient forward progress; sampling between the two yields the best results by balancing exploitation and exploration.
Second, Table~\ref{tab:aggregation} shows that simple aggregation schemes such as \emph{Best-of-N} (selecting a single plan) and \emph{Mandatory Selection} (enforcing all drafted plans) yield limited gains, whereas treating draft candidates as \emph{references} rather than strict constraints significantly improves accuracy.
Third, Table~\ref{tab:selection} shows that other selection rules (Random and Forecast-Direct) underperform, while the proposed adaptive selection achieves the highest accuracy.

\section{Conclusion}
\label{sec:conclusion}

In this work, we explore idle time, arising naturally from tool calls, as an underexploited computational resource to improve agent performance with minimal latency overhead.
Specifically, we introduce \methodname{}, which employs two complementary drafting strategies during tool execution. By adaptively selecting between these strategies and aggregating drafts once observations arrive, \methodname{} effectively utilizes idle time under observation uncertainty.
Extensive experiments across diverse agentic benchmarks demonstrate that \methodname{} generalizes to a wide range of tool-augmented scenarios.
We hope this work draws attention to idle-time utilization as a promising direction for building more capable LLM agents with minimal latency overhead.

\bibliography{main}
\bibliographystyle{unsrtnat}

\appendix
\newpage
\appendix
\onecolumn
\section{Additional Experimental Details} 
\label{sec:appendix_details}

\subsection{Datasets}
\label{sec:appendix_datasets}
We use three benchmarks that span complementary types of agentic execution: tool-using question answering (GAIA), multi-hop search (FRAMES), and machine-learning engineering (MLE-Bench). Below we describe the protocol for each.

\paragraph{GAIA and FRAMES.}
For GAIA, we evaluate on the full validation split across Levels 1--3, comprising 53, 86, and 26 tasks for Level~1, Level~2, and Level~3, respectively (165 tasks in total). For FRAMES~\cite{krishna2024frames}, a multi-hop question answering benchmark in which agents are repeatedly invoked as tools to perform sequential search-based reasoning, we use the first 50 samples for evaluation; this setup induces non-trivial idle periods on the orchestrator side. We use the same agent stack and tool budgets across all methods. For open-source models (Qwen3.5-4B and Gemma4-E4B), inference is served through 16 vLLM replicas on NVIDIA A6000 GPUs behind a load-balancing proxy; for Gemini-2.5-Flash, we use the official Vertex AI API.

\paragraph{MLE-Bench.}
We evaluate on MLE-Bench Lite~\cite{chan2024mle-bench}, i.e., the benchmark's Low-complexity split, which consists of 22 competitions. We follow the standard MLE-Bench Lite protocol with a 24-hour wall-clock budget per task for all methods. Each run starts from the same initial task state and follows the benchmark's standard interaction protocol. For every task, we report (i) whether the agent produces a submission (\emph{Made Submission}), (ii) whether the submission is accepted as valid by the benchmark (\emph{Valid Submission}), and (iii) the achieved leaderboard tier (\emph{Above Median}, \emph{Bronze}, \emph{Silver}, \emph{Gold}, and \emph{Any Medal}), computed using MLE-Bench's evaluation rules. All methods use Gemini-2.5-Flash as the agent backbone (via the Vertex AI API), the same budget and stopping criteria, and an identical software stack: a single GPU (NVIDIA RTX 3090) and 4 CPUs are provisioned to the agent for code execution and local tool use, with dependency versions held fixed across methods.

\subsection{Architectures}
\label{sec:appendix_model_settings}
Across all experiments, we consider Gemini-2.5-Flash\footnote{\url{https://docs.cloud.google.com/vertex-ai/generative-ai/docs/models/gemini/2-5-flash}}, Qwen3.5-4B\footnote{\url{https://huggingface.co/Qwen/Qwen3.5-4B}}, or Gemma4-E4B\footnote{\url{https://huggingface.co/google/gemma-4-E4B}}.
For agent frameworks, we use OAgents~\citep{zhu2025oagents} for GAIA 
and FRAMES, which provides a standard tool suite including text and 
audio processing, web browsing, and specialized sub-agents such as a 
dedicated web agent. For MLE-Bench, we use OpenHands~\citep{wang2025openhands},
which supports code generation and execution in interactive environments.

\subsection{Evaluation Protocol} By default, we use standard decoding with temperature $0.6$ and top-$p$ $1.0$. For Gemma4-E4B, we instead follow the recommended configuration on its official model card: temperature $1.0$, top-$p$ $0.95$, and top-$k$ $64$.
For GAIA and FRAMES, we report the mean across three random seeds for every (method, model) configuration. For MLE-Bench, we report results from a single seed since each run already spans the standard 24-hour budget per task, which provides sufficient stability.

\subsection{Implementation Details}
\label{sec:appendix_implementation_details}
\label{sec:appendix_prompts}
\label{sec:appendix_idlespec}

\paragraph{\methodname.}
We cap the number of candidate plans retained per idle window at $K{=}5$ across all experiments to prevent context-window overflow and keep the aggregation cost bounded; When a tool call returns before a draft finishes generating, 
the in-flight generation is interrupted and the partial draft is discarded. 
\methodname uses four prompts: a progressive drafting prompt (Figure~\ref{fig:prompt_progressive}); a recovery drafting prompt (Figure~\ref{fig:prompt_recovery}); a forecast prompt (Figure~\ref{fig:prompt_forecast}); an aggregation prompt (Figure~\ref{fig:prompt_aggregation}). All prompt templates are collected in Appendix~\ref{sec:appendix_prompt_templates}.

\paragraph{Baselines.}
\label{sec:appendix_baselines}
We compare \methodname against two baselines: \emph{Sequential Revision} (Figure~\ref{fig:prompt_seqrev}), which reflects on the latest observation to propose the next step, and \emph{Sleep-Time Compute} (Figure~\ref{fig:prompt_sleep_time}), which performs free-form inference over
a partially revealed problem during idle periods. We use the prompt released in the original Sleep-Time Compute repository\footnote{\url{https://github.com/letta-ai/sleep-time-compute}}, adapted from its original math-oriented template via a minimal modification: we apply the instruction
\textit{``The original prompt is designed for mathematical problem solving. Please minimally adapt it to better support \{task\}.''}, where \texttt{\{task\}} is replaced with \emph{``general agent problem''} for GAIA and \emph{``machine learning engineering''} for MLE-Bench. This adaptation introduces only small wording changes (e.g., replacing math-specific terms such as ``calculations'' with task-relevant terms like ``commands'' in MLE-Bench), while preserving the overall structure of the original prompt. A sensitivity analysis confirming that this baseline is robust to such phrasing variations is provided in Appendix~\ref{sec:appendix_prompt_sensitivity}.

\subsection{Algorithm Pseudocode}
\label{sec:appendix_algorithm}
\begin{algorithm}[H]
\caption{IdleSpec: Exploiting Idle Time via Speculative Planning for LLM Agents}
\label{alg:idlespec}
\begin{algorithmic}[1]
\Require Task $\mathcal{T}$, initial state $s_0$, language model $\mathcal{M}$, candidate cap $K$
\State Initialize Beta posterior $(\alpha, \beta) \leftarrow (1, 1)$ \Comment{uniform prior over $p = \Pr(\ell{=}\textsc{Prog})$}
\State Initialize empty draft buffer $\mathcal{D} \leftarrow \emptyset$
\State $s \leftarrow s_0$
\While{task not finished}
    \State Send $(s, \mathcal{D})$ to Main API and obtain next action
    \State Clear draft buffer $\mathcal{D} \leftarrow \emptyset$
    \State Execute the returned action
    \While{action execution in progress} \Comment{Drafting phase (in idle time)}
        \State Sample preference $\hat{p} \sim \mathrm{Beta}(\alpha, \beta)$
        \State Select drafting strategy
        $\mathbf{s}^\ast \leftarrow \textsc{Prog}$ if $\hat{p} > 0.5$, else $\textsc{Rec}$
        \If{$\mathbf{s}^\ast = \textsc{Prog}$}
            \State Generate progressive draft under observation uncertainty
        \Else
            \State Generate recovery draft conditioned on the current action step
        \EndIf
        \If{$|\mathcal{D}| < K$} \State Append generated draft to $\mathcal{D}$ \EndIf
    \EndWhile

    \State Receive observation $o$
    \State Ignore any in-flight idle draft responses and finalize $\mathcal{D}$
    \Comment{Forecast phase}
    \State Send $(\mathcal{T}, s, o)$ to Forecast API and obtain $\ell \in \{\textsc{Prog}, \textsc{Rec}\}$
    \State Update posterior $(\alpha, \beta)$ with $\ell$ using Eq.~\eqref{eq:posterior-update}
    \State $s \leftarrow s \,\Vert\, (\text{action}, o)$ \Comment{append step to trajectory}
\EndWhile
\end{algorithmic}
\end{algorithm}

\subsection{Broader Impact}
\label{sec:appendix_broader_impact}
\methodname{} reframes idle time during tool execution as a usable computational 
resource. On the positive side, it improves the achievable accuracy of agent 
systems within a fixed wall-clock budget without modifying the underlying model, 
which can lower the latency cost of capable agent execution and improve 
accessibility for latency-sensitive deployments.

On the negative side, the accuracy gains come from additional compute spent 
during the idle window (Appendix~\ref{sec:appendix_limitations}). Although 
end-to-end latency is unchanged, the underlying hardware is not free during 
idle periods, so this overhead translates into higher aggregate energy 
consumption and carbon footprint per task at scale.

\subsection{Limitations}
\label{sec:appendix_limitations}
While \methodname{} demonstrates consistent gains across diverse agentic scenarios, 
its effectiveness depends on the presence of sufficient idle time during tool 
execution. When tool calls return faster than a single LLM reasoning step, 
\methodname{} falls back to the vanilla baseline. In addition, although 
\methodname{} adds only minimal end-to-end latency overhead, the speculative 
drafting consumes additional LLM tokens during the idle window, which translates 
into higher per-task compute and monetary cost when running on metered APIs. 
For deployments where token cost rather than latency is the binding constraint, 
this trade-off should be considered explicitly.

\subsection{Future Work}
\label{sec:appendix_future_work}
We identify several promising directions for future work. A particularly 
compelling avenue is to decouple idle-time computation from the primary 
proprietary model by leveraging a small, specialized local model during 
idle periods. This setup may substantially reduce API inference cost and 
energy consumption while preserving the performance gains of the proprietary 
model on the critical path. More elaborate scheduling policies for idle time 
are also an interesting direction, including dynamically allocating speculative 
budgets based on task uncertainty or difficulty. Furthermore, extending 
\methodname{} beyond ReAct-style frameworks remains an open direction, 
including complex multi-agent paradigms and asynchronous or parallel 
tool-calling settings. Even when multiple tool calls are issued concurrently, 
the agent typically remains idle until the slowest call returns, and reasoning 
chains with sequential dependencies---such as multi-hop retrieval in GAIA or 
iterative code-debugging in MLE-Bench---can rarely be fully parallelized.

\section{Additional Analysis}
\label{sec:appendix_additional_analysis}

\subsection{Idle-Time Utilization vs.\ Accuracy}
\label{sec:appendix_idle_utilization}

\paragraph{Definition.}
For a task (episode) with $K$ tool-call steps, let $t^{(k)}_{\text{idle}}$ denote the tool execution duration of step $k$ (during which the agent is otherwise idle), and let $t^{(k)}_{\text{LLM-on-idle}} \le t^{(k)}_{\text{idle}}$ denote the portion of that interval actually spent on LLM computation (e.g., issuing requests, waiting on responses, and consuming generated drafts). We define the per-task \emph{idle-time utilization} as
\begin{equation}
\mathrm{ITU}_{\text{task}} = \frac{\sum_{k=1}^{K} t^{(k)}_{\text{LLM-on-idle}}}{\sum_{k=1}^{K} t^{(k)}_{\text{idle}}} \in [0, 1],
\label{eq:itu}
\end{equation}
so that a method that performs no idle-time computation has $\mathrm{ITU}_{\text{task}} = 0$, while perfect overlap of LLM work with the idle interval yields $\mathrm{ITU}_{\text{task}} = 1$. We report the average of $\mathrm{ITU}_{\text{task}}$ over all tasks (and seeds, when applicable) for each (method, benchmark) configuration.

\begin{figure}[t]
\centering
\begin{minipage}[c]{0.59\linewidth}
\centering
\includegraphics[width=0.75\linewidth]{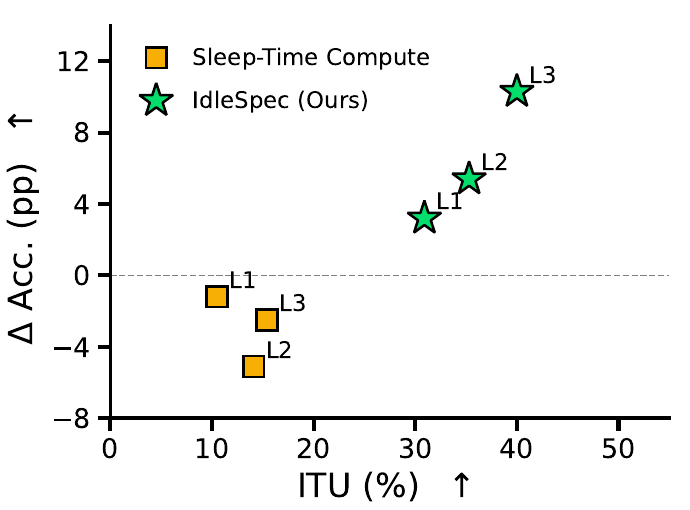}
\captionof{figure}{\textbf{Idle-Time Utilization (ITU) vs.\ Accuracy Gain ($\Delta$).} GAIA with Qwen3.5-4B, one point per difficulty level.}
\label{fig:gaia-scatter}
\end{minipage}\hfill
\begin{minipage}[c]{0.4\linewidth}
\centering
\captionof{table}{\textbf{Ultra-Short Idle-Time Ratio.} Average fraction of tool calls per task whose execution time is shorter than a single LLM reasoning step on GAIA. The reasoning-step duration used as the threshold is measured from Gemini-2.5-Flash.}
\label{tab:ultra-short-ratio}
\vspace{0.6em}
\begin{tabular}{cc}
\toprule
Level & Ultra-short ratio \\
\midrule
1 & 0.25 \\
2 & 0.26 \\
3 & 0.27 \\
\bottomrule
\end{tabular}
\end{minipage}
\end{figure}

Figure~\ref{fig:gaia-scatter} illustrates the relationship between idle-time utilization and the resulting accuracy gain over the vanilla baseline on GAIA with Qwen3.5-4B. \methodname{} reaches 34.6\% idle-time utilization on average and turns it into consistent positive gains across all three difficulty levels (up to +10.3 points on Level~3), whereas Sleep-Time Compute utilizes only 13.2\% of the available idle time and even degrades accuracy at every level. This indicates that performance improvements are not solely driven by how much idle time is used but also by how it is exploited, and reflects the fact that Sleep-Time Compute does not account for variation in tool-call durations.

\subsection{Ultra-Short Idle-Time Ratio}
\label{sec:appendix_idle_length}
The body's analysis of idle-time length (Section~\ref{sec:ablation}) bins
GAIA samples by their \emph{ultra-short ratio}: the fraction of a
sample's tool calls whose execution time is shorter than a single LLM
reasoning step. Table~\ref{tab:ultra-short-ratio} reports the average
ultra-short ratio per GAIA difficulty level (25--27\% across Levels 1--3),
confirming that most tool calls leave usable idle time for speculative
drafting. The accuracy impact of binning by this ratio is reported in
the body as Table~\ref{tab:ultra-short-impact}.

\subsection{Extended Efficiency Analysis}
\label{sec:appendix_extended_cost}
We conduct an extended cost analysis on the full GAIA dataset with Gemini-2.5-Flash to characterize the wall-clock and token cost of \methodname relative to the vanilla baseline. To obtain stable per-task latency measurements, all numbers in this table come from a single fixed seed run in an isolated environment (no co-located workloads, fixed concurrency, identical tool stack across methods). Table~\ref{tab:extended_efficiency} reports the results across the three GAIA difficulty levels: \methodname matches the vanilla baseline in end-to-end latency at every level, while delivering consistent accuracy gains (up to +7.7 points on Level~3).

Like other test-time scaling approaches, \methodname spends additional computation to improve task performance. The key distinction is \emph{where} this computation is placed: by scheduling drafting onto otherwise wasted idle periods rather than serializing it on the critical path, \methodname turns idle time into accuracy gains with only minimal end-to-end latency overhead, attaining accuracy comparable to test-time scaling methods while substantially shortening wall-clock time (Table~\ref{tab:token-analysis}).

\begin{table}[h]
\centering
\small
\caption{\textbf{Extended Efficiency Analysis on the Full GAIA Dataset with Gemini-2.5-Flash} (single fixed seed, isolated environment). Per-task wall-clock latency in seconds. \methodname matches the vanilla baseline in latency while consistently improving accuracy across all three difficulty levels.}
\label{tab:extended_efficiency}
\vspace{0.5em}
\begin{tabular}{llcc}
    \toprule
    Level & Method & Accuracy (\%) & Latency (s) \\
    \midrule
    \multirow{2}{*}{Level 1} & Vanilla        & 64.2 & 107 \\
                             & + \methodname  & 66.0 & 99 \\
    \midrule
    \multirow{2}{*}{Level 2} & Vanilla        & 41.9 & 191 \\
                             & + \methodname  & 46.5 & 196 \\
    \midrule
    \multirow{2}{*}{Level 3} & Vanilla        & 26.9 & 374 \\
                             & + \methodname  & 34.6 & 376 \\
    \bottomrule
\end{tabular}
\end{table}

\begin{table}[h]
\centering
\small
\begin{minipage}[t]{0.5\linewidth}
\centering
\caption{\textbf{Generalization to Other Agent Frameworks.} Accuracy (\%) on the full GAIA dataset with Gemini-2.5-Flash, when integrating \methodname into SmolAgents.}
\label{tab:framework_generalization}
\vspace{0.8em}
\begin{tabular}{lcccc}
    \toprule
    Method & Level 1 & Level 2 & Level 3 & Avg. \\
    \midrule
    SmolAgents             & 60.4 & 45.3 & 19.2 & 41.6 \\
    \rowcolor{green!10}
    + \textbf{\methodname}           & \textbf{68.0} & \textbf{47.8} & \textbf{30.8} & \textbf{48.9} \\
    \bottomrule
\end{tabular}
\end{minipage}\hfill
\begin{minipage}[t]{0.45\linewidth}
\centering
\caption{\textbf{Sensitivity Analysis of Prompt Design} (single fixed seed). GAIA average accuracy (\%) of the Sleep-Time Compute under three prompt variants on Gemini-2.5-Flash.}
\label{tab:prompt_sensitivity}
\vspace{0.4em}
\begin{tabular}{lc}
    \toprule
    Prompt Variant & Average. \\
    \midrule
    Version 1 (Original) & 44.5 \\
    Version 2            & 44.0 \\
    Version 3            & 44.0 \\
    \bottomrule
\end{tabular}
\end{minipage}
\end{table}

\subsection{Generalization to Other Agent Frameworks}
\label{sec:appendix_framework_generalization}
To assess whether the benefits of \methodname extend beyond a single agent implementation, we employ it in another agent stack, SmolAgents\footnote{\url{https://github.com/huggingface/smolagents}}, a widely used framework designed for flexibility and minimalism. We re-implement the drafting, forecast, and aggregation modules within \texttt{smolagents}'s execution loop without changing any prompts or hyperparameters, and evaluate the resulting agent on the full GAIA dataset with Gemini-2.5-Flash. Table~\ref{tab:framework_generalization} shows that \methodname improves accuracy across all three difficulty levels in this stack, raising average accuracy from 41.6\% to 48.9\% (+7.3 points) and yielding the largest gain on Level~3 (+11.6 points). The pattern of improvement closely mirrors that observed on the OAgents stack, indicating that the gains arise from the general principle of overlapping speculative reasoning with idle time rather than from any framework-specific implementation choice.

\subsection{Sensitivity Analysis of Prompt Design}
\label{sec:appendix_prompt_sensitivity}
Because the Sleep-Time Compute baseline relies on a prompt adapted from a math-oriented template (Appendix~\ref{sec:appendix_baselines}), we run a sensitivity analysis to confirm that our reported numbers do not hinge on the particular adaptation we use. We construct three prompt variants by re-running the same minimal-adaptation procedure with different model families (Claude and Gemini) as the rewriter, and re-evaluate the Sleep-Time Compute baseline under each. Table~\ref{tab:prompt_sensitivity} shows that the GAIA average stays within a 0.5-point band across the three variants (44.0--44.5\%), confirming that the comparison against Sleep-Time Compute reported in the main paper is robust to the choice of prompt adaptation.

\subsection{Qualitative Analysis}
\label{sec:appendix_qualitative}

We present two FRAMES tasks (Qwen3.5-4B, seed~0) on which \emph{both} Vanilla and Sleep-Time Compute fail while \methodname{} succeeds. For each task we summarize the trajectory of every method in compact form (tool-call sketch, key observation, final answer) and highlight where Vanilla and Sleep-Time Compute diverge from the correct chain.

\paragraph{Example 1: \texttt{frames\_3} (Compositional Sports--Temporal Reasoning).}
\textit{``As of August~1, 2024, which country were holders of the FIFA World Cup the last time the UEFA Champions League was won by a club from London?''} \textbf{Gold answer:} \emph{France}. The required chain is: (i) most recent London-club Champions League winner $\rightarrow$ Chelsea, May~2021; (ii) FIFA World Cup holder \emph{on that date} $\rightarrow$ France, because France won the 2018 tournament and remained the reigning holder until Argentina won in December~2022 --- i.e., on Chelsea's May~2021 victory date, the trophy was still held by France.

\begin{tcolorbox}[colback=red!4, colframe=red!60!black, title=\textsc{Vanilla} (Pred: Argentina / GT: France), fontupper=\small, breakable]
\begin{enumerate}
  \item \texttt{search\_agent("List of UEFA Champions League winners by London clubs, years won, most recent victory before 2024")} $\rightarrow$ ``Chelsea FC \dots {} most recent victory \textbf{2021}.''
  \item \texttt{search\_agent("Which country won the FIFA World Cup in 2021?")} $\rightarrow$ ``No country won the FIFA World Cup in 2021 \dots{} the 2022 winner was Argentina.''
  \item \texttt{final\_answer("Argentina")}.
\end{enumerate}
\textbf{Failure Mode.} Vanilla collapses ``World Cup holder \emph{at the time of} the 2021 Champions League final'' into ``World Cup winner \emph{of} 2021'' and is then misled by the search agent's correction (``\dots{} the 2022 winner was Argentina''), committing to that without re-deriving who actually held the trophy on the date of the Chelsea win.
\end{tcolorbox}

\begin{tcolorbox}[colback=orange!5, colframe=orange!70!black, title=\textsc{Sleep-Time Compute} (Pred: England / GT: France), fontupper=\small, breakable]
\begin{enumerate}
  \item[1-4.] Four interchangeable list-the-winners queries (\texttt{search\_agent}, \texttt{web\_search}, \texttt{wiki}); all return empty observations.
  \item[5.] \texttt{search\_agent("Which London clubs have won the UEFA Champions League and in which years")} $\rightarrow$ ``Chelsea: 2012, 2021. Manchester United: 1968, 1999, 2008.''
  \item[6.] \texttt{final\_answer("England")}.
\end{enumerate}
The pre-computed inference produced by the sleep-time worker before step~6 is itself the source of the wrong commit:
\begin{quote}\small
\textit{``Chelsea won the Champions League in 2012 and 2021 (specifically May 29, 2021). [\dots] Chelsea is a club located in London, England. Therefore, the country associated with the winning club is England (or United Kingdom). [\dots] \textbf{The country is England.}''}
\end{quote}
\textbf{Failure Mode.} The sleep-time pre-computation reframes the task from ``country that held the World Cup at the time of the 2021 Chelsea win'' into ``country in which the winning club is located,'' and concludes the wrong country in advance. The executor, anchored to that pre-computed answer, never fires the World Cup hop and submits ``England'' --- the failure is committed inside the sleep-time context, not in the executor's tool use.
\end{tcolorbox}

\begin{tcolorbox}[colback=blue!4, colframe=blue!60!black, title=\textsc{\methodname{}} (Pred: France / GT: France), fontupper=\small, breakable]
\begin{enumerate}
  \item \texttt{search\_agent("When did Chelsea FC last win the UEFA Champions League?")} $\rightarrow$ ``Chelsea FC last won the UEFA Champions League in \textbf{2021}.''
  \item \texttt{search\_agent("Who won the FIFA World Cup in 2022?")} $\rightarrow$ ``Argentina won the 2022 FIFA World Cup\dots{} December~18, 2022.''
\end{enumerate}
At this idle window (after the 2022 observation), the sleep-time worker drafted two progressive and three recovery candidate plans in parallel. Two recovery plans flagged the exact failure that Vanilla and Sleep-Time Compute fall into:
\begin{quote}\small
\textbf{Recovery~\#0.} \textit{``Re-evaluate the question's intent: determine if `holders of the FIFA World Cup' refers to the winner of the World Cup \emph{in that specific year} (which would be none) or \textbf{the winner of the World Cup \emph{ending on or before that date} (i.e., the 2018 tournament)}. Search for the FIFA World Cup winner of the last World Cup prior to or on the date of the 2021 Champions League victory to see which interpretation yields a valid answer.''}\\[2pt]
\textbf{Recovery~\#2.} \textit{``Search for the FIFA World Cup winner of the tournament that \textbf{concluded prior to August~1, 2024} [\dots] \textbf{specifically identifying the 2018 winner (France)} to confirm the correct `holder' status relative to the 2021 Champions League victory.''}
\end{quote}
The actual observation matches both recovery assumptions (the 2022 winner does not satisfy ``holder at the time of the 2021 Chelsea win''), so the executor adopts that gate in its next steps:
\begin{enumerate}
\setcounter{enumi}{2}
  \item \texttt{search\_agent("Who is the current holder of the FIFA World Cup as of August~1, 2024?")} $\rightarrow$ ``Argentina is the reigning FIFA World Cup holder\dots'' (verifies that ``current'' holder $\ne$ holder \emph{at the time of} the 2021 Chelsea win).
  \item \texttt{search\_agent("Who won the 2018 FIFA World Cup?")} $\rightarrow$ ``France'' (executes the recovery plan).
  \item \texttt{final\_answer("France")}.
\end{enumerate}
\textbf{Why This Works.} The injected recovery plans encode the missing semantic gate --- ``holder \emph{at the time of} 2021, not winner \emph{of} 2021'' --- and the executor's next thought adopts the cross-check before committing. Vanilla collapses this gate (commits to Argentina from a 2022 hop), and Sleep-Time Compute never reaches the World Cup hop at all because its pre-computed inference already concluded the answer was the country of the club.
\end{tcolorbox}

\paragraph{Example 2: \texttt{frames\_25} (Date Arithmetic).}
\textit{``What was the age difference between Mike Tyson and Tyson Fury on the respective days on which they lost their first ever fights? Represent the figure in years only.''} \textbf{Gold answer:} \emph{12 years}.

\begin{tcolorbox}[colback=red!4, colframe=red!60!black, title=\textsc{Vanilla} (Pred: 12.74 / GT: 12), fontupper=\small, breakable]
\begin{enumerate}
\setcounter{enumi}{0}
  \item[1-4.] Search for both fighters' first-loss dates and birthdates: Tyson ($1966$-$06$-$30$, lost $1990$-$02$-$11$), Fury ($1988$-$08$-$12$, lost $2024$-$05$-$18$).
  \item[5-6.] Compute fractional ages: Tyson $23.619$~years, Fury $36.356$~years.
  \item[7.] \texttt{final\_answer(round(12.736481861738536, 2))} $\rightarrow$ \texttt{12.74}.
\end{enumerate}
\textbf{Failure Mode.} Reasoning is correct but the answer is reported as a fractional year rather than an integer-year difference; the question explicitly says ``represent the figure in years only,'' which the judge interprets as integer years. Vanilla never re-checks the formatting requirement before committing.
\end{tcolorbox}

\begin{tcolorbox}[colback=orange!5, colframe=orange!70!black, title=\textsc{Sleep-Time Compute} (Pred: 11 / GT: 12), fontupper=\small, breakable]
\begin{enumerate}
  \item[1-8.] Tool calls retrieve Tyson's first-loss date (Feb 11, 1990) and various candidate Fury loss dates from search.
\end{enumerate}
The pre-computed inference produced by the sleep-time worker before the final action is the source of the wrong answer:
\begin{quote}\small
\textit{``\textbf{Mike Tyson's Birth Date:} March~6, 1966. \dots{} \textbf{Tyson Fury's First Loss Date:} Confirmed as \textbf{March~24, 2023} (vs.\ Dmitry Bivol). Note: The previous search results mentioned May~18, 2024 (vs.\ Usyk), but that was the rematch or a subsequent loss. \dots{} \textbf{Age:} 23 years \dots{} \textbf{Age:} 34 years \dots{} Age\_Fury (34) - Age\_Tyson (23) = \textbf{11 years}.''}
\end{quote}
\textbf{Failure Mode.} The sleep-time pre-computation hallucinates two facts. (i) Mike Tyson's birthdate is reported as March~6, 1966 --- the actual date is June~30, 1966; the correct value was returned by an earlier search but is silently overwritten in the sleep-time context. (ii) Tyson Fury never lost to Bivol; the May~18, 2024 Usyk fight was actually his first professional loss, but the sleep-time worker reasons it away as ``a rematch or subsequent loss'' to fit a fabricated 2023 Bivol date. The executor inherits both errors and the pre-computed final difference of 11 years, and submits 11 verbatim.
\end{tcolorbox}

\begin{tcolorbox}[colback=blue!4, colframe=blue!60!black, title=\textsc{\methodname{}} (Pred: 12 / GT: 12), fontupper=\small, breakable]
\begin{enumerate}
  \item[1-7.] Tool calls retrieve and verify both first-loss dates (Tyson Feb 11, 1990; Fury May 18, 2024) and both birthdates (Tyson June 30, 1966; Fury August 12, 1988).
\end{enumerate}
At this idle window, the sleep-time worker drafts plans that explicitly fix the arithmetic template and the formatting requirement:
\begin{quote}\small
\textbf{Progressive~\#0.} \textit{``Once the verification confirms May~18, 2024 as the first loss date, calculate the exact ages of both fighters on those respective dates using \textbf{their verified birth dates (June~30, 1966 for Tyson; August~12, 1988 for Fury)}. Compute the difference between the two ages and \textbf{provide the final answer in years only}, ensuring the calculation accounts [for whether the birthday has passed].''}
\end{quote}
The plan pins three things that the other two methods miss: (a) the correct Tyson birthdate (June~30, not the hallucinated March~6 used by Sleep-Time Compute), (b) a symmetric ``compute integer age on each loss date'' template, and (c) the explicit ``in years only'' formatting requirement.
\begin{enumerate}
\setcounter{enumi}{7}
  \item Final code applies the template verbatim, prints integer ages 22 and 34, and submits \texttt{final\_answer(12)}.
\end{enumerate}
\textbf{Why This Works.} The injected plan turns the failure modes of the other two methods into pinned constraints in the executor's context: the pre-verified birthdates close the door on Sleep-Time Compute's hallucination, and the explicit ``years only'' instruction closes the door on Vanilla's float-precision formatting. Neither gate is created mid-step; both come from plans drafted in the prior idle window.
\end{tcolorbox}

\paragraph{Example 3: GAIA \texttt{56137764} (Pivoting to a Different Retrieval Strategy).}
\textit{``Which contributor to the version of OpenCV where support was added for the Mask-RCNN model has the same name as a former Chinese head of government when the names are transliterated to the Latin alphabet?''} \textbf{Gold answer:} \emph{Li Peng} (Premier of the PRC, 1987--1998). The required chain is: (i) identify the OpenCV release that added Mask-RCNN $\rightarrow$ \emph{4.0.0}; (ii) retrieve the \emph{full contributor list} for that release; (iii) retrieve the list of former Chinese premiers/presidents in standard Pinyin; (iv) cross-reference the two lists. The trap is that a casual reading of the question pulls the agent toward ``find \emph{the} commit author of the Mask-RCNN PR,'' which is a strict subset of the contributor list and which does not contain the gold answer.

\begin{tcolorbox}[colback=red!4, colframe=red!60!black, title={\textsc{Vanilla} (Pred: Vadim Pisarevsky / GT: Li Peng; 16 steps)}, fontupper=\small, breakable]
\begin{enumerate}
  \item[1-3.] \texttt{search\_agent} for ``OpenCV Mask-RCNN commit author'' returns ``Dmitry Lychak.'' The agent commits to the framing ``find the single PR/commit author whose name matches a Chinese premier.''
  \item[4-13] Eleven further \texttt{search\_agent} calls, each a slight variant of ``OpenCV Mask-RCNN commit author / PR / release-notes,'' return mutually inconsistent names: ``Sunita Nayak,'' ``Alexander Mordvintsev,'' ``Tim M.,'' ``dkurt (Dmitry Kurtaev),'' ``Dmitry Khitrov.'' The Thought field repeatedly says ``\textit{conflicting information \dots{} let me try a different approach},'' but each ``different approach'' is another commit-author search.
  \item[14-16.] The agent gives up on the search and falls back to prior knowledge, submitting \texttt{final\_answer("Vadim Pisarevsky")} --- a well-known OpenCV co-founder, but never a Chinese premier.
\end{enumerate}
\textbf{Failure Mode.} Vanilla locks in the wrong task framing on step~1 (``find the commit author'') and never widens the retrieval to the full contributor set. Each of the 11 follow-up searches reuses the same template; the conflicting answers are read as ``the search agent is unreliable'' rather than as a signal that the framing itself is wrong, so the agent stays inside the same approach until step exhaustion forces a hallucinated commit.
\end{tcolorbox}

\begin{tcolorbox}[colback=orange!5, colframe=orange!70!black, title={\textsc{Sleep-Time Compute} (Pred: Li Peng / GT: Li Peng; 9 steps)}, fontupper=\small, breakable]
\begin{enumerate}
  \item \texttt{search\_agent("OpenCV Mask-RCNN support added version contributor")} $\rightarrow$ partial release-note information.
\end{enumerate}
The pre-computed sleep-time inference at this point speculatively enumerates plausible Chinese premier names alongside OpenCV-related names, producing a long ``\textit{maybe the matching contributor is one of: \textbf{Li Peng}, Li Keqiang, Zhou Enlai, \dots}'' paragraph. The executor copies that enumeration directly into the next query:
\begin{enumerate}
\setcounter{enumi}{1}
  \item[2-8.] \texttt{search\_agent("OpenCV Mask-RCNN contributor Li Keqiang Li Peng Zhou Enlai")} and several variants. After enough such queries, ``Li Peng'' is confirmed to appear in the OpenCV 4.0.0 contributor list.
\setcounter{enumi}{8}
  \item \texttt{final\_answer("Li Peng")}.
\end{enumerate}
\textbf{Mode.} Sleep-Time Compute happens to land on the gold answer in 2 of 3 seeds because its pre-inference dumps a list of candidate Chinese names into context and the executor gets to verify them by name-matching. The mechanism is search-by-guessing rather than a structured pivot: across seeds, the same prompt sometimes lists the right names, sometimes does not (seed~2 returns ``Unable to determine'' after 16 steps).
\end{tcolorbox}

\begin{tcolorbox}[colback=blue!4, colframe=blue!60!black, title={\textsc{\methodname{}} (Pred: Li Peng / GT: Li Peng; 5 steps)}, fontupper=\small, breakable]
\begin{enumerate}
  \item \texttt{search\_agent("OpenCV version added Mask-RCNN model support")} $\rightarrow$ ``Mask-RCNN was added in OpenCV \textbf{4.0.0}.''
\end{enumerate}
At the idle window during step~1's search execution, the sleep-time worker frames the next sub-goal as ``retrieve the contributor list of that release'' rather than ``find the Mask-RCNN commit author'' --- so the alternative framing that traps Vanilla is never even drafted. The selected plan reads:
\begin{quote}\small
\textbf{Progressive~\#0.} \textit{``Once the search agent returns the specific OpenCV version number, the next sub-goal is to \textbf{search for the list of contributors or authors specifically for that release version}. After obtaining the contributor list, verify that the data includes names in a standard transliterated format (e.g., Pinyin) to facilitate comparison with former Chinese heads of government before proceeding to the cross-referencing step.''}
\end{quote}
This is the single decisive intervention of the run: it flips the retrieval scope from ``the commit author'' (Vanilla's permanent trap) to ``the full contributor set,'' before the executor commits to its second action. The executor follows the plan exactly:
\begin{enumerate}
\setcounter{enumi}{1}
  \item \texttt{search\_agent("OpenCV 4.0.0 contributors list")} $\rightarrow$ broad list including ``Alexander Alekhin, Dmitry Kurtaev, Hamdi Sahloul, Maksim Shabunin, \dots, \textbf{Li Peng}, \dots''
\end{enumerate}
\begin{enumerate}
\setcounter{enumi}{2}
  \item \texttt{search\_agent("list of former Chinese prime ministers premier names transliterated to Latin alphabet")} $\rightarrow$ Pinyin list.
\end{enumerate}
At the next idle window (during step~3's search), the selected plan closes the task with an explicit string-match-then-commit gate, and even pre-enumerates the right candidate name:
\begin{quote}\small
\textbf{Progressive~\#3.} \textit{``Perform a case-insensitive string match between these names and the OpenCV 4.0.0 contributor list, \textbf{specifically checking names like ``Li Peng''}, ``Wu Zhiwen,'' ``Kuang Fangjun.'' If a match is found, verify it against the original contributor list, then \textbf{use \texttt{final\_answer} to provide the matching name}.''}
\end{quote}
\begin{enumerate}
\setcounter{enumi}{3}
  \item Inline cross-reference between the two lists prints ``Li Peng'' as the unique match.
  \item \texttt{final\_answer("Li Peng")}.
\end{enumerate}
\textbf{Why This Works.} The decisive contribution is the first idle window's plan, which widens retrieval scope before the second action is chosen --- once the executor's step~2 query is ``contributors list of 4.0.0'' instead of ``commit author of the Mask-RCNN PR,'' the cross-reference and commit steps follow on a short straight path. The next two idle windows then narrow the corridor (cross-reference axis, then commit gate) so no post-evidence loop is possible. Across all three seeds \methodname{} converges to ``Li Peng'' (3/3); Sleep-Time Compute converges by guessing and is correct only 2/3, while Vanilla never escapes the commit-author framing (0/3).
\end{tcolorbox}

\clearpage
\section{Prompt Templates}
\label{sec:appendix_prompt_templates}

\begin{figure}[H]
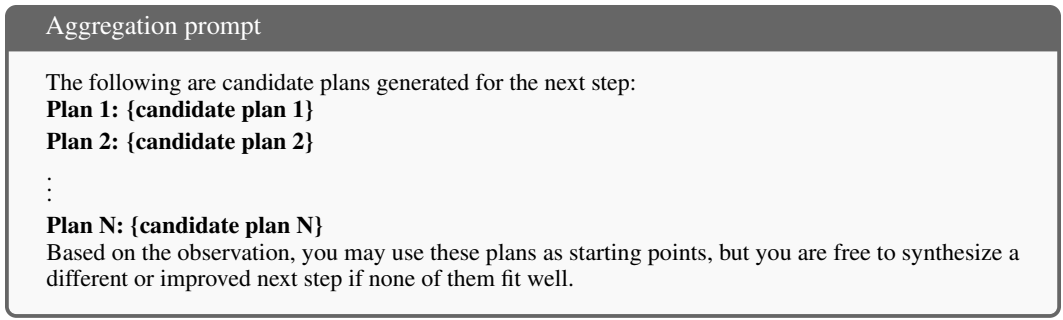

\begin{tcolorbox}[colback=gray!5, colframe=black!60, title=Aggregation prompt, fontupper=\small, width=\linewidth]
The following are candidate plans generated for the next step:

\textbf{Plan 1:} \textbf{\{candidate plan 1\}} \\[2pt]
\textbf{Plan 2:} \textbf{\{candidate plan 2\}} \\[2pt]
$\vdots$ \\[2pt]
\textbf{Plan N:} \textbf{\{candidate plan N\}}

Based on the observation, you may use these plans as starting points, but you are free to synthesize a different or improved next step if none of them fit well.
\end{tcolorbox}
\caption{Aggregation prompt that consumes the candidate plans together with the just-arrived observation and produces the final next step.}
\label{fig:prompt_aggregation}
\end{figure}

\begin{figure}[H]
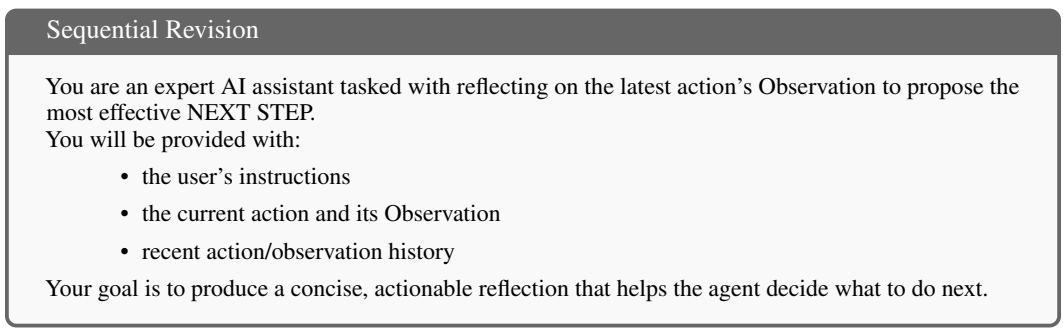

\begin{tcolorbox}[colback=gray!5, colframe=black!60, title=Sequential Revision, fontupper=\small, width=\linewidth]
You are an expert AI assistant tasked with reflecting on the latest action's Observation to propose the most effective NEXT STEP.

You will be provided with:
\begin{itemize}
\item the user's instructions
\item the current action and its Observation
\item recent action/observation history
\end{itemize}
Your goal is to produce a concise, actionable reflection that helps the agent decide what to do next.
\end{tcolorbox}
\caption{Sequential Revision prompt. The model reflects on the executed action and its observation to propose the next step.}
\label{fig:prompt_seqrev}
\end{figure}

\begin{figure}[H]
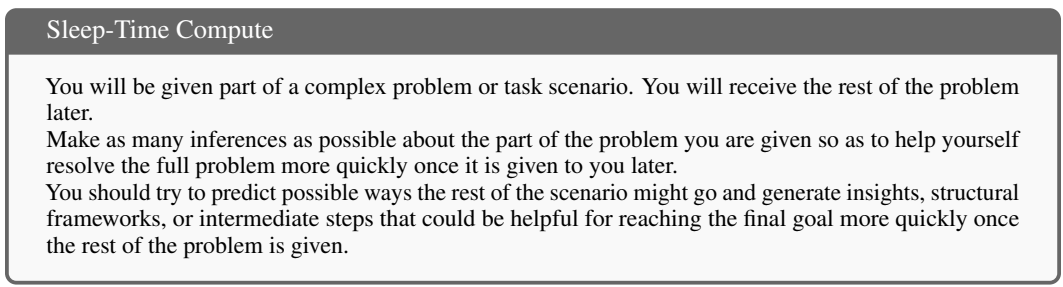

\begin{tcolorbox}[colback=gray!5, colframe=black!60, title=Sleep-Time Compute, fontupper=\small, width=\linewidth]
You will be given part of a complex problem or task scenario. You will receive the rest of the problem later.

Make as many inferences as possible about the part of the problem you are given so as to help yourself resolve the full problem more quickly once it is given to you later.

You should try to predict possible ways the rest of the scenario might go and generate insights, structural frameworks, or intermediate steps that could be helpful for reaching the final goal more quickly once the rest of the problem is given.
\end{tcolorbox}
\caption{Sleep-Time Compute prompt. The model is asked to pre-compute inferences and intermediate insights from a partially revealed problem during idle periods, which can later be reused once the full problem becomes available.}
\label{fig:prompt_sleep_time}
\end{figure}

We provide here the full text of the prompts referenced in Appendix~\ref{sec:appendix_implementation_details}. Figures~\ref{fig:prompt_progressive}--\ref{fig:prompt_aggregation} list the four prompts used by \methodname (progressive drafting, recovery drafting, forecast, aggregation), and Figures~\ref{fig:prompt_seqrev}--\ref{fig:prompt_sleep_time} list the prompts used by the Sequential Revision and Sleep-Time Compute baselines.

\begin{figure}[H]
\begin{tcolorbox}[colback=gray!5, colframe=black!60, title=Progressive drafting prompt, fontupper=\small, width=\linewidth]
You are an expert AI assistant tasked with generating the most effective and efficient NEXT ACTION STEP to take after the current action step completes and its observation becomes available.

The current action step is running in parallel.
You will not see the observation yet; assume it will arrive and will be used to decide the next step.

If plan\_history is non-empty, target a different sub-goal from those already proposed. \\[2pt]
\textbf{Plan history:} [\{prior draft 1\}, \ldots, \{prior draft K\}]
\end{tcolorbox}
\caption{Progressive drafting prompt used by \methodname to speculatively generate the next action while the current tool call is in flight.}
\label{fig:prompt_progressive}
\end{figure}

\begin{figure}[H]
\begin{tcolorbox}[colback=gray!5, colframe=black!60, title=Recovery drafting prompt, fontupper=\small, width=\linewidth]
You are an expert AI assistant tasked with generating a RECOVERY strategy while the current action step is executing.

Assume that the current action step fails to make progress or gets stuck.
Propose EXACTLY one distinct recovery plan that takes a different approach from what has already been tried.

If plan\_history is non-empty, choose a meaningfully different angle --- different evidence source, different verification strategy, or different decomposition. \\[2pt]
\textbf{Plan history:} [\{prior draft 1\}, \ldots, \{prior draft K\}]
\end{tcolorbox}
\caption{Recovery drafting prompt used by \methodname to draft an alternative plan that diverges from the current trajectory in case the in-flight step fails.}
\label{fig:prompt_recovery}
\end{figure}

\begin{figure}[H]
\begin{tcolorbox}[colback=gray!5, colframe=black!60, title=Forecast prompt, fontupper=\small, width=\linewidth]
You are a forecast module that evaluates whether the agent's current action step is making forward progress toward solving the Task, and decides which strategy the agent should take next.

The goal is to assess the current progress state and output a single decision: whether the agent should continue with a PROGRESSIVE strategy or switch to a RECOVERY strategy.

STRATEGY SEMANTICS:
\begin{itemize}
\item \textbf{PROGRESSIVE:} The agent is making meaningful forward progress. Assume the current trajectory is valid and focus on exploitation by continuing or refining the existing plan with minimal deviation.
\item \textbf{RECOVERY:} The agent appears stalled, misguided, or unproductive. Assume the current trajectory may be flawed and encourage exploration by deviating from the existing plan, considering alternative approaches or corrective actions not previously explored.
\end{itemize}
\end{tcolorbox}
\caption{Forecast prompt used by \methodname after the observation arrives to choose between the progressive and recovery candidates.}
\label{fig:prompt_forecast}
\end{figure}

\end{document}